# Emotion Correlation Mining Through Deep Learning Models on Natural Language Text

Xinzhi Wang, Luyao Kou, Vijayan Sugumaran, Xiangfeng Luo, and Hui Zhang

*Abstract*—Emotion analysis has been attracting researchers' attention. Most previous works in the artificial-intelligence field focus on recognizing emotion rather than mining the reason why emotions are not or wrongly recognized. The correlation among emotions contributes to the failure of emotion recognition. In this article, we try to fill the gap between emotion recognition and emotion correlation mining through natural language text from Web news. The correlation among emotions, expressed as the confusion and evolution of emotion, is primarily caused by human emotion cognitive bias. To mine emotion correlation from emotion recognition through text, three kinds of features and two deep neural-network models are presented. The emotion confusion law is extracted through an orthogonal basis. The emotion evolution law is evaluated from three perspectives: one-step shift, limited-step shifts, and shortest path transfer. The method is validated using three datasets: 1) the titles; 2) the bodies; and 3) the comments of news articles, covering both objective and subjective texts in varying lengths (long and short). The experimental results show that in subjective comments, emotions are easily mistaken as anger. Comments tend to arouse emotion circulations of love–anger and sadness–anger. In objective news, it is easy to recognize text emotion as love and cause fear–joy circulation. These findings could provide insights for applications regarding affective interaction, such as network public sentiment, social media communication, and human–computer interaction.

## I. Introduction

EMOTION is complex, individualized, subjective, and sensitive to context. Emotion guides decision, prepares the body for action, and shapes the ongoing behavior [1]. Philosophers tend to conclude that emotion is a subjective response to the objective world, which means emotion stems from the interaction between society and individuals. Individual emotion is complex in at least the following three aspects.

1) Steady individual value is formed through long-term experience. Emotion response among individuals differs even in the same context. For instance, the Napoleonic War is disputed with two opposite views. The supporters hold the view that the war attacks the French feudal force and prompts historical progress, while the opponents think the war is unjustified due to its aggressive purpose.
2) Misunderstanding occurs when individuals communicate. The understanding of the context varies as individual prior backgrounds differ. The opinion of an individual becomes more profound when obtaining more knowledge on target events. The misunderstanding of initial emotion happens when there is a prior knowledge gap between the information sender and the receiver.
3) Individual emotion turbulence exists. The turbulence is affected by external instant negative or positive mood. Emotion changes along with instant conditions for the same event. For most individuals, it is a common phenomenon in daily life that external conditions influence internal emotions. For example, a sweet-sounding tweet can also be disturbing when one's work performance is judged negatively.

On the one hand, the emotion of individuals is complex due to individualized long-term social experiences, interpersonal misunderstandings, and external instant mood influence. On the other hand, public emotion concerning the social event is complex because of the following reasons.

1) The social event, with a set of aspects, is complex. The information about social events is released online, including all kinds of topics. Individualized netizens tend to pay attention to different aspects. The emotion of the readers may be diverse, giving different aspects of

Manuscript received July 12, 2019; revised December 1, 2019 and February 12, 2020; accepted March 26, 2020. This work was supported in part by the National Key Research and Development Program of China under Grant 2017YFC0803300, in part by the National Science Foundation of China under Grant 91646201, Grant U1633203, and Grant 91746203, in part by the High-Tech Discipline Construction Fundings for Universities in Beijing (Safety Science and Engineering), and in part by the Beijing Key Laboratory of City Integrated Emergency Response Science. This article was recommended by Associate Editor S. Ozawa. *(Xinzhi Wang and Luyao Kou contributed equally to this work.) (Corresponding author: Xinzhi Wang.)*

Xinzhi Wang and Xiangfeng Luo are with the School of Computer Engineering and Science, Shanghai University, Shanghai 200444, China, and also with the Shanghai Institute for Advanced Communication and Data Science, Shanghai University, Shanghai 200444, China (e-mail: wxz2017@shu.edu.cn; luoxf@shu.edu.cn).

Luyao Kou and Hui Zhang are with the Institute of Public Safety Research, Department of Engineering Physics, Tsinghua University, Beijing 100084, China, and also with the Beijing Key Laboratory of City Integrated Emergency Response Science, Tsinghua University, Beijing 100084, China (e-mail: kly18@mails.tsinghua.edu.cn; zhhui@mail.tsinghua.edu.cn).

Vijayan Sugumaran is with the Department of Decision and Information Sciences, Oakland University, Rochester, MI 48309 USA, and also with the Center for Data Science and Big Data Analytics, Oakland University, Rochester, MI 48309 USA (e-mail: sugumara@oakland.edu).

the same social event. If the reader is concerned about more than one aspect, then his/her words may carry more than one kind of emotion.
2) Public emotion is compound and diverse. There is still controversy on the classification of emotions in the research area of social psychology. One of the well-acknowledged categories is dividing human emotion into six categories, that is, love, joy, anger, sadness, fear, and surprise. Shaver *et al.* [2] employed a tree structure to describe these basic emotions. Ekman's theory, which was similar to Shaver's theory, included the controversial emotion surprise [3]. This article classifies emotions into these six categories. The proposed method is also applicable when given other kinds of emotion categories.

Emotions are correlated rather than independent, which contributes to the complexity of individual and public emotions. Emotion correlation mining can help analyze the individual and public emotions at least in the following applications:
1) *Public Sentiment Analysis:* As Zhao *et al.* [4] pointed out, emotion variation contributes a lot to netizens' behavior comprehension and abnormal event detection in social media.
2) *Social Media Communication:* It is beneficial to generate low ambiguous messages that are empathetic to the information receiver for both news compilation and interpersonal communication. Emotion correlation mining can provide clues for the expression of the intended emotion.
3) *Human–Computer Interaction (HCI):* Emotions contribute to improve HCI, for example, social companion robots. Emotion is intuitive in providing robots clues to understand and predict behavior for humanistic reaction.

The potential applications of emotion analysis have been attracting a lot of attention from researchers. However, most efforts have focused on emotion recognition while neglecting emotion correlation mining. This article tries to fill the gap between emotion recognition and emotion correlation mining. We propose an approach for emotion correlation mining, which includes extracting emotion confusion and evolution. The confusion of emotion refers to the distance among emotions, obtained through projecting the emotions to a space with appropriate dimensions. The evolution of emotion is analyzed to shed light on the developing direction of the network public sentiment. The mining process is based on the emotion recognition results of the deep learning classification models using text.

This article is organized as follows. Related works are reviewed in Section II. Section III presents the term definition, the potential error causes, and directions to mine emotion correlation. Sections IV and V provide a detailed description of the calculation method for emotion confusion and evolution, respectively. In Section VI, the experimental evaluation of the proposed approach is described. Finally, conclusions are drawn in Section VII.

## II. Related Work

Emotion analysis has been drawing researchers' attention in recent years. Text emotion distribution learning [5], [6], considered as one of the most important research areas, contributes to many applications. The research on emotion recognition [7], [8] is opening up numerous opportunities pertaining to social media in terms of understanding users' preferences, habits, and their contents. The application of the subjective and emotional data from social media includes but is not limited to sentiment analysis [9], [10]; sarcasm detection [11]; event dissemination [12]; user clustering [13], [14]; and user behavior analysis [15]. Some tasks of the applications are combined together, such as multitask assignment on sentiment classification and sarcasm detection, which employs the deep neural network in natural language processing (NLP) tasks [16].

Emotion analysis from text is one of the hot topics in modern natural language understanding. Embedding and attention mechanisms help a lot with emotion recognition in deep learning methods. Continuous word representations, including word2vec [17], weighted word embedding [18], and the derivatives [19] denoted words with dense embeddings and provided new ideas for automatic feature mining. Later, different kinds of attention mechanisms and pretrained models were proposed. Wang *et al.* [20] proposed an embedded recursive neural network for improving emotion recognition. Barros *et al.* [21] introduced a personalized affective memory. In 2017, Vaswani *et al.* [22] proposed a new network architecture, the Transformer, based solely on attention mechanisms, dispensing with recurrence and convolutions entirely. With the capability of modeling bidirectional contexts, denoising autoencoding-based pretraining, such as BERT [23], and an autoregressive pretraining method, such as XLNet [24], achieved better performance than many other pretrained models on GLUE tasks. Some researchers solved the emotion recognition task through graph modeling [25], such as a capsule network. Popular natural language understanding models, such as long short-term memory (LSTM) [26], convolutional neural network (CNN) [27], recursive autoencoders [28], adversarial learning [29], and attention mechanism [30], have been applied for emotion analysis and classification tasks. Electroencephalography signals and facial expression sequences were also used for emotion recognition with deep learning models [31], [32]. More complex and classification-oriented deep learning models made it harder to understand the correlation among emotions even with remarkable recognition accuracy.

Emotion analysis, as an important traditional branch of knowledge mining, is categorized into three levels, namely: word level, sentence level, and document level. In word level, emotion words were extracted mainly through three ways: 1) manual approach [33]; 2) dictionary-based approach [34]; and 3) corpus-based approach [35]. Strapparava and Valitutti [36] developed WordNet affect through tagging a subset of synets with affective meanings in English WordNet (EWN). Staiano and Guerini [37] presented DepecheMood, an emotion lexicon produced automatically by harvesting social media data annotated with emotion scores. Then, Badaro *et al.* [38] provided EmoWordNet by expanding DepecheMood with the synonymy semantic relation from EWN. Emotion lexicons for different languages were developed. In SemEval 2018 Task 1: Affect in Tweets [39], labeled data from English, Arabic, and Spanish tweets are created for each task. Badaro et al. [40] achieved the best result in the SemEval 2018 emotion classification subtask for the Arabic language. Features that they used were word embeddings from AraVec, and emotion features extracted from ArSEL [41] and NRC emotion lexicon. In the sentence-level analysis, intrasentential and intersentential emotion consistency were explored [42]. Qiu et al. [43] employed dependency grammar to describe relations for double propagation between features and opinions. Ganapathibhotla and Liu [44] adopted dependency grammar for the emotion analysis of comparative sentences. The conditional random fields (CRFs) method [45] was used as the sequence learning technique for extraction. A multitask multilabel (MTML) classification model was proposed to classify sentiment and topics concurrently [46]. By doing this, the closely related tasks, that is, sentiment and topic classification have been improved. Machine-learning methods were widely used in both the sentence and document levels. Naive Bayesian [47], maximum entropy classification [48], graphical model [49], and pattern recognition methods [50] were employed frequently. Zhao et al. [51] explored the correlations among different microblogs for social event detection. Hu and Flaxman [52] provided multimodel sentiment analysis by combining visual analysis and NLP to predict various emotional states of the user in social media. Most of the previous works focused on recognizing emotions from text rather than why emotion was wrongly recognized.

Studies on emotion or sentiment propagation provided clues on building correlation emotion. The common phenomena of spread of sentiment [53] have been found, including positive [54] and negative sentiments [55]. Stieglitz and Dang-Xuan [56] explored the association between emotion and the user's information-sharing behavior. They found that emotionally charged Twitter messages tend to be retweeted with both higher quantity and speed in the social media setting. Some findings showed that negative sentiment may contribute to content diffusion more than positive sentiment in the news domains, such as the findings of Hansen et al. [57]. Fan et al. [58] divided the sentiment into four categories differing from the previously oversimplified sentiment classification (e.g., polarity detection [59]) and revealed that anger was more likely to spread than joy, disgust, and sadness especially toward social problems via Weibo in China. In other non-news domains, the opposite conclusion may even hold due to the complexity of emotions. Ferrara and Yang [53] explored the dynamics of emotional contagion with a random sample of Twitter users. They pointed out that users were more likely to adopt positive emotions than negative emotions on Twitter. The propagation prediction of emotion or sentiment was built on the precise emotion or sentiment recognition, which still neglected mining the correlation of emotions.

All the above works improved the performance of emotion recognition. However, just as Wilson et al. [60] pointed out, a single text may contain multiple opinions. Parrott [61] demonstrated that human emotions were prototyped and complex. Most of the recent works just focus on recognizing the emotion expressed in text and emotion diffusion in social media. Little attention is paid to associate emotion calculation [62] based on the quantitative and engineering approach with interemotion correlation typically studied in psychology. Analysis of correlation among emotions caused by the complexity of emotion is few focused and few covered in the literature in computer science.

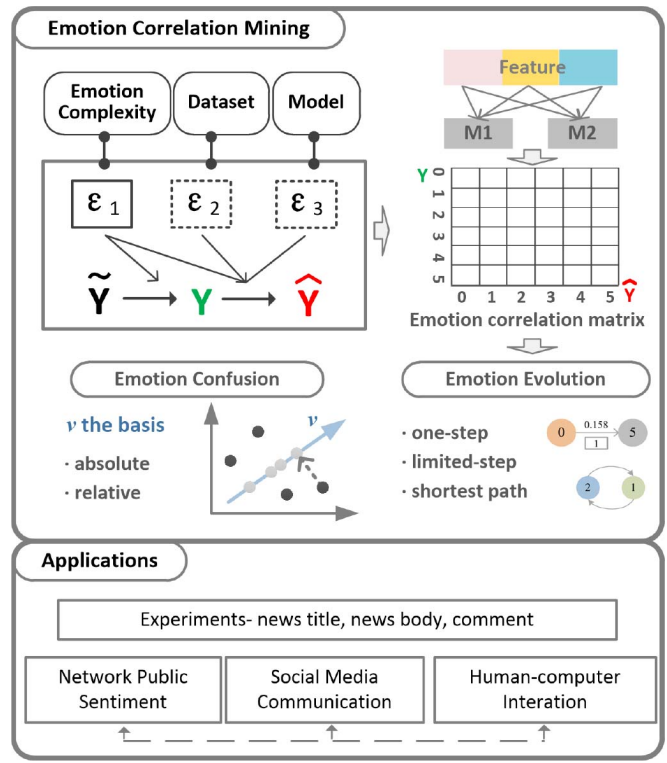

Fig. 1.    Framework of emotion correlation mining. 0: love, 1: fear, 2: joy, 3: sadness, 4: surprise, 5: anger.

## III. MODEL FRAMEWORK

This article aims to mine the potential and meaningful correlation among emotions from Web news. The emotion classification models in Section III-C focus on discriminating the emotion orientation of input texts. Three features of datasets and two deep neural-network models are proposed to do that. The calculation result supports emotion correlation mining in Sections IV and V. The total framework is depicted in Fig. 1.

### A. Term Definitions

1) The terms "human emotion," "public emotion," and "social event" are often used in affective computing studies. While definitions of these terms may appear self-explanatory. In this article, they are defined as follows.
   a) Emotion is defined as a strong feeling deriving from one's circumstances, mood, or relationships with others in Oxford Dictionary [63]. Emotion responses to significant internal and external event [64]. This article divides human emotion into six categories as presented by Shaver et al. [2].

    b) Public emotion refers to the sum of the individuals' emotional states [65]. The emotions of audiences may differ when they pay attention to different aspects of the same event. These various emotions constitute public emotion.

    c) Social event indicates all kinds of events that are published online. Common topics of a social event include health, government, education, business, entertainment, unusual events, etc.

2) Emotion correlation is described as emotion confusion and evolution. Several laws are concluded.

    a) The confusion of emotion refers to "distance" among emotions. Absolute confusion of an emotion denotes the average probability measured by the distance that the emotion is confused with all other emotions. Relative confusion of the emotion is based on the relative distance. For example, if the distance between fear and surprise is shorter than that between fear and anger, it reveals that in terms of fear, the relative confusion degree of surprise is higher than that of anger. Fear and surprise are more likely to be confused. Basically, there is only one absolute confusion value for a given emotion, but multiple relative confusion values for that given emotion, one with each other emotion.

    b) The evolution of emotion refers to the emotion changes during the process of event propagation. Misjudgment of emotions is one of the important factors in the evolution of emotion. For example, a reader recognizes a text which contains love as joy, and then another netizen mistakes the text which contains joy as surprise. The above misjudgments contribute to the evolution of love–joy–surprise.

    c) The laws mentioned in this article stem from correlations among emotions. The interemotion correlation is summarized as several laws. The confusion and the evolution of emotion correspond to the emotion confusion law discussed in Section IV and the emotion evolution law in Section V, respectively. The emotion evolution law includes a misjudgment law of emotion and a circulation law of emotion.

### B. Error Analysis

As mentioned in the introduction, both emotions and social events are complex. Consequently, the essence of public emotions may cause calculation errors. Dataset and emotion classification models also contribute to the overall errors.

*1) Errors Caused by Emotion Complexity:* The complexity is composed of two parts, that is, the diversity of emotions and the complexity of social events. The complexity of the problem will inevitably lead to errors without considering other factors. Generally, the emotion of individuals is complex and is impacted by individualized long-term social experiences, misunderstandings caused by prior background knowledge, and external instant mood influences. Furthermore, public emotion is compound and diverse. Social events are complex, which are characterized by a set of aspects. The complexity of an event has a positive correlation with the number of aspects. Hence, if a reviewer pays attention to more than one aspect, then his/her words may carry more than one kind of emotion.

*2) Errors Caused by Dataset:* Text features, which represent text, are abstract basic language units. Characters (letters) form a single word, and words make up phrases, then phrases build sentences, paragraphs, and chapters. Character and word (explicit) are two features used frequently. In this article, implicit expression of words is employed as the third feature to reduce the sparsity caused by using words. These three kinds of features capture text information from different levels.

  1) *Character:* Letter in language. The basic feature of the text, whose number is limited and thus compact in the corpus.

  2) *Implicit Expression:* Synonym tag of words from the synonymous dictionary. If several words are synonyms, they share the same synonym tag, extracted from the HIT synonymous dictionary [HIT IR-Lab Tongyici Cilin (Extended)].

  3) *Explicit Expression:* Word in language. The number of features in explicit expression is not less than that in implicit expression, as synonyms share the same tag in implicit expression. In a sufficiently large corpus, there are more words than synonyms and more synonyms than characters.

Taking "love" and "like" for example, the character features are six letters "l," "o," "v," "e," "i," and "k." In implicit expression, the feature is the synonym tag for the reason that the two words are synonymous. It means that the synonym tag extracted from the HIT dictionary is employed to substitute the original two words. In explicit expression, the features are love and like, the semantics of which are dispersed compared to the implicit expression. Character, implicit, and explicit features carry distinguishable representative capability. Accordingly, the three features, which differ in efficiency and precision, are used to reduce the errors caused by the dataset.

*3) Errors Caused by Models:* Misunderstandings often occur when humans try to figure out the emotion from the text. Similarly, the emotion classification results of models may differ, given the same input text as models' inner calculation logic varies. Considering the models as intelligent agents, models will make mistakes. To reduce the error caused by models, this article presents more than one model. Through the complementary advantages of the models, the results can be of higher authenticity and reliability. Besides, the decisions of the majority make sense even when the majority makes wrong decisions sometimes. We try to find the mistakes that most models make and try to conclude the common mistake laws. Two deep learning models are established to minimize the errors caused by a single model.

*4) Error Influence:* The above three errors account for the deviation from the emotion correlation. Let the errors caused by emotion complexity, datasets, and models be denoted as $\varepsilon_1$, $\varepsilon_2$, and $\varepsilon_3$, respectively. Suppose $\widetilde{Y}$ is the ground truth of the emotion, which should be obtained from text writers and is therefore unavailable. $Y$ is marked by the public and is used to substitute $\widetilde{Y}$. The details are shown in Section VI-A.

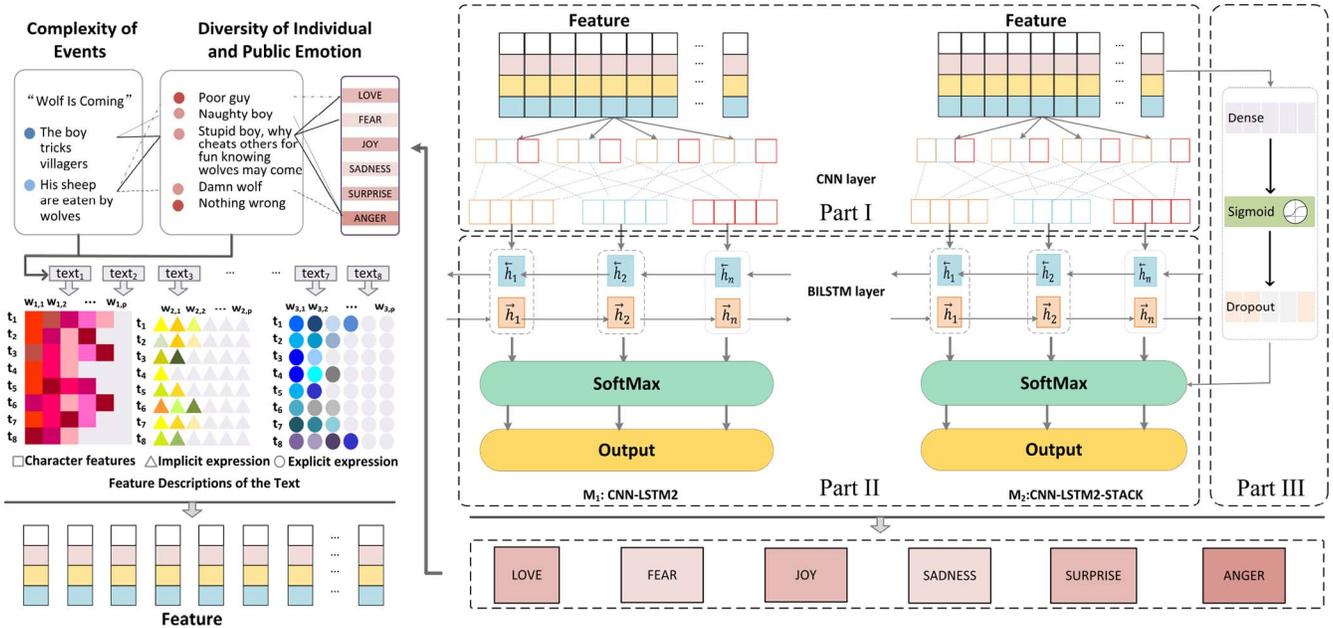

Fig. 2. Measures taken to minimize the errors of emotion classification and interemotion correlation.

$Y$ indicates the emotion label of the text, which acts as the input label for the deep learning models. $\widehat{Y}$ represents the label predicted by the models. Optimally, the three values should be equal, $\widetilde{Y} = Y = \widehat{Y}$. However, it is almost impossible to keep the consistency of the ground truth of emotion, the publicly annotated and the model-predicted labels in all cases as shown in Fig. 1. On the one hand, emotion complexity $\varepsilon_1$ leads to the deviation between the public-annotated label $Y$ and the ground truth $\widetilde{Y}$, formulated as $\varepsilon_1 \propto \sigma_1 = [(\mathbf{Y}-\widetilde{\mathbf{Y}})/(\widetilde{\mathbf{Y}})]$. On the other hand, the three types of errors contribute to the deviation between the model-predicted label $\widehat{Y}$ and the public-annotated label $Y$, that is, $\varepsilon_1, \varepsilon_2, \varepsilon_3 \propto \sigma_2 = [(\widehat{\mathbf{Y}}-\mathbf{Y})/(\mathbf{Y})]$.

$\varepsilon_1$ stems from the human emotion cognitive bias and the complexity of social events, which is irreducible. Errors caused by datasets ($\varepsilon_2$) and models ($\varepsilon_3$) are considered to be minimized, as they are reducible. To do that, the voting mechanism is presented by using more than one model and by importing three kinds of input features, namely, character features, implicit expression, and explicit expression. The details are shown in Fig. 2. The above measures are taken to estimate the confusion and evolution of target public emotion. By combining three features and two emotion classification models, this article tries to minimize the errors caused by the dataset ($\varepsilon_2$) and the model ($\varepsilon_3$).

### C. Emotion Classification Model

Here, two deep neural-network models, CNN-LSTM2 (M1) and CNN-LSTM2-STACK (M2), are employed for emotion recognition. In both models, the length of an input text can be either short or long. The output of the models is one of the six kinds of emotions, that is, love, joy, anger, sadness, fear, and surprise. The calculation process can be divided into three parts. M1 is constructed with Part I and Part II. M2 is constructed by adding an additional Part III to M1. The brief details of the three parts are described as follows. More details can be found in Appendix A.

*1) Part I (Feature Processing):* Part I focuses on feature processing which transforms the original features into dense vector information. There are four operations: vector lookup; window sliding; convolutional calculation; and rectified linear units (ReLUs) activation.

*2) Part II (Emotion Calculation):* Part II focuses on the emotion calculation. There are five operations, that is, LSTM calculation, dropout operation, average calculation, fully connected layer, and softmax. Recurrent neural networks (RNNs) are powerful for sequence processing tasks, such as text classification. As a variant of RNNs, an LSTM network is capable of classifying and predicting sequence when there are very long time lags of unknown size between important time steps.

*3) Part III (Original Feature Attention):* Part I and Part II construct the model M1. However, with the neural network going deep, the backward fine-tuning process in M1 becomes weak, and the vanishing gradient problem occurs. To solve this problem, a second model M2 is constructed by associating M1 with Part III.

The other operations in Part II remain the same. Part III aims to emphasize the impact of the input feature embedding on the emotion calculation result. In other words, by stacking Part III, the network pays more attention to the original feature information. The two models M1 and M2 use deep neural-network methods to calculate text emotion. The detailed description is demonstrated by Wang et al. [62]. In our model design, both long and short texts can act as inputs. Three parts are introduced, including feature processing, emotion calculation, and original feature attention. The details are shown in Fig. 2. The emotion calculation results of the two models with three text features support the interemotion correlation analysis.

## IV. Confusion of Emotion

Let $x_0(s_i|s_j, F_\alpha, M_\beta)$ represent the number of the texts in a dataset with an input label ($Y = s_j$) and an output label ($\widehat{Y} = s_i$), given a feature $F_\alpha$ and a model $M_\beta$. Let $\mathbf{x}_0^{(k)} \in R^{6\times 6}$ be a matrix whose element is $x_0(s_i|s_j, F_\alpha, M_\beta)$. $k$ is an integer ranging from 1 to 6, with $\alpha$ being an integer ranging from 1 to 3, and $\beta$ being an integer ranging from 1 to 2.

Then, the emotion correlation matrix $\mathbf{x}_1^{(k)}$ is obtained by applying normalization to each row of $\mathbf{x}_0^{(k)}$. This matrix $\mathbf{x}_1^{(k)} \in R^{6\times 6}$ comprises the proportion that emotions are recognized correctly and mistaken as other five emotions

$$\mathbf{x}_1^{(k)} = \text{NORMALIZATION}\left(\mathbf{x}_0^{(k)}\right). \tag{1}$$

There are $m_1$ (six) kinds of feature–model combinations of choices in total. That means, for each dataset, there are $m_1$ corresponding emotion correlation matrices

$$\mathbf{X}_1 = \left\{\mathbf{x}_1^{(k)} | k = 1, 2, \ldots, m_1\right\} \tag{2}$$

where $\mathbf{X}_1$ is the set of matrices for the dataset $X$.

### A. Emotion Principal Component Analysis

An emotion feature extraction method is applied on the emotion correlation matrix $\mathbf{x}_1^{(k)}$. Consider $\mathbf{x}_1^{(k)} = [\mathbf{x}_{1,1}^{(k)}, \ldots, \mathbf{x}_{1,m_0}^{(k)}]$, where $\mathbf{x}_{1,i}^{(k)} \in \mathbb{R}^D$ is a column vector with dimensionality $D$. This method focuses on finding out an orthonormal basis with dimensionality $D_1$ ($D_1 < D$), onto which the variance of the projected data is maximized. The orthonormal basis, which is the most distinct direction of the space with dimensionality $D_1$, stands for the emotions features.

The orthonormal basis is denoted as $\mathbf{v} = [\mathbf{v}_1, \ldots, \mathbf{v}_{D_1}]$, $\mathbf{v}_i \in \mathbf{v}$, where $\mathbf{v}_i \in \mathbb{R}^D$ is a $D$-dimensional vector. Let $\mathbf{v}_i^T\mathbf{v}_j = \delta$ $\forall i, j = 1, \ldots, D_1$, which means $\mathbf{v}$ is a standard orthonormal basis. Each data point $\mathbf{x}_{1,n}^{(k)}$ is projected onto $\mathbf{v}_1$ as $\mathbf{v}_1^T \mathbf{x}_{1,n}^{(k)}$. $\mathbf{v}_1^T \bar{\mathbf{x}}_1^{(k)}$ denotes the mean of the projected data, where $\bar{\mathbf{x}}_1^{(k)}$ is the column vector mean. The variance of the projected data is calculated by the following equation:

$$\frac{1}{m_0} \sum_{n=1}^{m_0} \left(\mathbf{v}_1^T \mathbf{x}_{1,n}^{(k)} - \mathbf{v}_1^T \bar{\mathbf{x}}_1^{(k)}\right)^2$$
$$= \mathbf{v}_1^T \left[\frac{1}{m_0} \sum_{n=1}^{m_0} \left(\mathbf{x}_{1,n}^{(k)} - \bar{\mathbf{x}}_1^{(k)}\right)\left(\mathbf{x}_{1,n}^{(k)} - \bar{\mathbf{x}}_1^{(k)}\right)^T\right] \mathbf{v}_1$$
$$= \mathbf{v}_1^T \Psi \mathbf{v}_1 \tag{3}$$

where $\Psi$ is the covariance matrix corresponding to $\mathbf{x}_1^{(k)}$. In order to obtain the maximal variance, that is, maximize $\mathbf{v}_1^T \Psi \mathbf{v}_1$ with respect to $\mathbf{v}_1$, the Lagrange multiplier is introduced and denoted as $\lambda_1$

$$\mathbf{v}_1^T \Psi \mathbf{v}_1 + \lambda_1 \left(1 - \mathbf{v}_1^T \mathbf{v}_1\right) \tag{4}$$

by setting the first derivative with respect to $\mathbf{v}_1$ as 0, the condition of the maximal variance is given by

$$\Psi \mathbf{v}_1 = \lambda_1 \mathbf{v}_1 \tag{5}$$

which indicates that $(\mathbf{v}_1, \lambda_1)$ is an eigenpair of $\Psi$. Left multiply by $\mathbf{v}_1^T$ and consider $\mathbf{v}_1^T \mathbf{v}_1 = 1$ which is the property of an orthonormal basis. The variance can be formulated as

$$\mathbf{v}_1^T \Psi \mathbf{v}_1 = \lambda_1. \tag{6}$$

The variance of the data projected onto $\mathbf{v}_1$ is the maximum when $\mathbf{v}_1$ is an eigenvector of $\Psi$ corresponding to the maximal eigenvalue $\lambda_1$. $\lambda_1$ is known as the first principal component in the principal component analysis (PCA) [66]. The additional directions that maximize the projected variance are obtained by choosing the new directions that are orthonormal to those already available. The eigenvectors of the covariance matrix, $[\mathbf{v}_1, \ldots, \mathbf{v}_{D_1}]$, are the most distinct directions of the matrix, which can be demonstrated by induction.

The above discussion is the derivation process based on the maximum variance theory. Moreover, the orthonormal basis also conforms to the minimum error theory that minimizes the average projected cost, which can be developed likewise.

A set of principal component vectors and corresponding weights are acquired through the emotion feature extraction method. The principal component vectors refer to a standard orthonormal basis, the calculation of which observes two principles: 1) project the primary data points onto the vectors such that the projected data variance is maximized and 2) the average distance between the primary data points and their projections is minimized. Principal component vectors represent the most distinct directions of the emotion correlation matrix. Emotion PCA is mainly used to obtain these vectors to represent the emotional features. Then, distances (confusion degree) between emotions can be calculated.

### B. Emotion Confusion Feature Extraction

Each emotion correlation matrix stands for a perspective to estimate emotion correlation. This section adds the matrix $\mathbf{x}_1^{(m_1+1)}$ equaling the mean of $\mathbf{x}_1^{(1)}, \ldots, \mathbf{x}_1^{(m_1)}$ to the matrix $\mathbf{X}_1$ in (2). Accordingly, there are $m_1 + 1$ points of view presented to explore the confusion of emotion.

Emotion is divided into $m_0$ (six) categories. For an emotion correlation matrix $\mathbf{x}_1^{(k)}$, the emotion feature is a set of principal component vectors $\mathbf{v} = [\mathbf{v}_0, \mathbf{v}_1, \ldots, \mathbf{v}_{m_0-1}]$ corresponding to the eigenvalues $\lambda = [\lambda_0, \lambda_1, \ldots, \lambda_{m_0-1}]$. In practice, the first $m_2$ vectors are adopted. The decrease of the dataset dimensionality leads to information loss. The cumulative variance contribution of the first $m_2$ vectors represents the proportion of retained information in the original dataset. Then, the weighted average of the first $m_2$ vectors is given by

$$\mathbf{v}^{(k)} = \frac{\sum_{i=0}^{m_2} \lambda_i \mathbf{v}_i}{\sum_{i=0}^{m_2} \lambda_i}. \tag{7}$$

The feature of the dataset with $m_1$ kinds of emotion correlation matrix is denoted as $\mathbf{X}_2 = [\mathbf{v}^{(1)}, \mathbf{v}^{(2)}, \ldots, \mathbf{v}^{(m_1+1)}]^T$ that can also be represented by the column vectors

$$\mathbf{X}_2 = [\mathbf{e}_0, \mathbf{e}_1, \ldots, \mathbf{e}_{m_0-1}] \in \mathbb{R}^{(m_1+1)m_0} \tag{8}$$

where $\mathbf{e}_g$ is the $g$th column vector, and $g$ ranges from 0 to $m_0 - 1$.

## C. Confusion Degree

Let $e_g$ indicate the feature of emotion $s_g$. The confusion degree refers to distance among emotions, which is obtained from subtraction of the corresponding columns. The closer distance represents a higher level of confusion. Subtract $e_0, \ldots, e_{m_0-1}$ from $e_g$. Then, the emotion $s_g$ centered distance matrix are obtained and represented as

$$\mathbf{X}_3^g = \left[ d_0^g, d_1^g, \ldots, d_{m_0-1}^g \right] \in \mathbb{R}^{(m_1+1)m_0} \quad (9)$$

where $d_i^g \in \mathbf{X}_3^g$ is a column vector. The elements of $d_i^g$ represent the distance between emotion $s_i$ and emotion $s_g$, from $m_1 + 1$ points of view. $d_g^g$ equals $\mathbf{0}$.

*1) Absolute Confusion Degree:* Absolute confusion degree represents the likelihood that an emotion is confused with other emotions. An increase in the confusion degree of emotion $s_g$ with any other emotions will enhance the absolute confusion degree of $s_g$

$$\mathbf{A}^g = \frac{\sum_{i=1}^{m_0} d_i^g}{(m_1+1)m_0} \quad (10)$$

where the larger $\mathbf{A}^g$ means a farther distance between $s_g$ and all other emotions, that is, lower absolute confusion degree of $s_g$.

*2) Relative Confusion Degree:* For a matrix $\mathbf{X}_3^g$, the six elements in each row are sorted in ascending order, that is, $\mathbf{X}_4^g = \text{ARGSORT}(\mathbf{X}_3^g)$. The sequence matrix $\mathbf{X}_4^g$ is given by

$$\mathbf{X}_4^g = \left[ se_0^g, se_1^g, \ldots, se_{m_0-1}^g \right] \in \mathbb{R}^{(m_1+1)m_0} \quad (11)$$

where the element of the first column $(se_0^g)$ indicates emotion $s_g$. The second column's $(se_1^g)$ element denotes the emotion that has the maximal confusion degree with emotion $s_g$, the first column' elements. Conversely, the last column's $(se_{m_0-1}^g)$ element, indicates the emotion that has the minimal confusion degree with emotion $s_g$, the first column' elements.

For the sequence matrix $\mathbf{X}_4^g$, if all the elements in a column are identical, it means that the $m_1 + 1$ perspectives are the same. A reliable result is obtained. That is, the more the identical number of the elements in a column, the more reliable the result will be. The introduction of information entropy eliminates untrustworthy results

$$\mathbf{H}(\mathbf{U}) = \mathbf{E}\left[ -\log p_i \right] = -\sum_{i=1}^{n} p_i \log p_i \quad (12)$$

where $p_i$ represents the proportion of a emotion in a column. Entropy indicates the degree of chaos. The lower information entropy refers to a higher consistency, that is, a more reliable result. Consequently, the columns of $\mathbf{X}_4^g$, whose information entropy is below the average, are extracted and analyzed.

## V. EVOLUTION OF EMOTION

Emotion confusion refers to the distance among emotions resulting from mutual misjudgment between two emotions. The confusion of emotion is bidirectional and, therefore, nondirectional. The evolution of emotion is directional, focusing on emotion shift during social events propagation. Emotion evolution can be evaluated from the following aspects.

$\mathbf{x}_1^{(k)}$ indicates the emotion correlation matrix mentioned in Section IV. Let $\text{prob}_n(s_i, s_j)$ represent the probability that $s_i$ is recognized as $s_j$ after $n$ shift steps. The corresponding path is denoted as $\text{trace}_n(s_i, s_j)$. Every shift step corresponds to a misjudgment between two emotions. $\text{prob}_n(s_i, s_j)$ is the multiplication of misjudgment probability of every shift step.

### A. One-Step Shift

The top misunderstanding emotion pair is figured out given an emotion $s_i$. According to the matrix $\mathbf{x}_1^{(k)}$, compare $\text{prob}_1(s_i, s_j)$ in terms of each emotion $s_j$. Then, the top pair is the two emotions with the maximal $\text{prob}_1(s_i, s_j)$. The trace is denoted as $\text{trace}_1(s_i, s_j) = s_i \to s_j$

$$\max\{\text{prob}_1(s_i, s_j) | j = 0, \ldots, m_0 - 1, j \neq i\}. \quad (13)$$

### B. Limited-Step Shifts

Under three conditions, that is, given initial emotion, given ultimate emotion, and given initial and ultimate emotion, the emotion evolution within limited steps is observed.

To extract the most possible shift path, the probability product $\text{prob}_n(s_i, s_j)$ is calculated and compared. However, the product may close to zero after $n$ shift steps. The logarithm of the matrix $\mathbf{x}_1^{(k)}$ is therefore employed. Specifically, the logarithm of the misjudgment probability for each shift step is summed and used, which is positively correlated with $\text{prob}_n(s_i, s_j)$.

Let the initial and ultimate emotions be denoted as $s_{\text{ini}}$ and $s_{\text{ult}}$, respectively. When $s_{\text{ini}}$ or $s_{\text{ult}}$ is given, there are $n^{m_0-1}$ kinds of potential shift paths in $n$ shift steps with the existence of misjudgments. When $s_{\text{ini}}$ and $s_{\text{ult}}$ are both given, there are $n^{m_0-2}$ kinds of potential shift paths. Under the above three conditions, the shift path with the maximal probability is extracted. In other words, at every shift step, the most possible emotion that is misjudged with the former emotion are selected for the emotion flow. The above process is expressed as follows:

$$\max\{\log \text{prob}_n(s_{\text{ini}}, s_j) | j = 0, \ldots, m_0 - 1\} \quad (14)$$
$$\max\{\log \text{prob}_n(s_i, s_{\text{ult}}) | i = 0, \ldots, m_0 - 1\} \quad (15)$$
$$\max\{\log \text{prob}_n(s_{\text{ini}}, s_{\text{ult}})\}. \quad (16)$$

### C. Shortest Path Transfer

Given the initial and ultimate emotion, the shortest path between two emotions is observed. The shortest path from emotion $s_{\text{ini}}$ to $s_{\text{ult}}$ is either one or $n$ steps through other emotions. Specifically, the trace with the maximal probability of misjudgments among one to $n$ shift steps is the shortest one

$$\max\{\max\{\log \text{prob}_n(s_{\text{ini}}, s_{\text{ult}})\} | n \in \mathbb{N}^*\}. \quad (17)$$

## VI. EXPERIMENTS

### A. Datasets

The datasets are collected from one of the most popular social network platforms, news channel (http://news.sina.com.cn/society/moodrank/). Each news item is split into three parts: 1) the news comment; 2) the news body; and 3) the news title, where the comment is treated as subjective text, and the news body and title are regarded as objective text. The emotion labels of the three datasets are generated through the vote of the public, strong rules, and manual selection. The public who read the news can vote for the news information on six emotions. First, only the data with total votes more than 200 are counted. Second, we label the data with emotion whose vote ratio count for more than 50% and discard data that do not meet the requirement. Take data with vote 50(love), 50(joy), 50(anger), 50(sadness), 50(fear), and 260(surprise) for instance, the data are labeled as surprise. If the vote of surprise is 240, the data are then discarded. Third, we check the data to see if the data are abnormal. The distribution of emotions in the three datasets is shown in Appendix B Table I. The comments contain more than 150 000 sample texts. The news bodies and news titles both contain more than 24 000 sample texts. The performance of the two models with the three features on the testing data, are illustrated in Appendix B Table II. The details of the models are analyzed by Wang *et al.* [62].

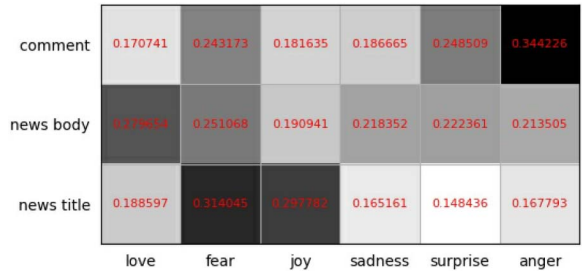

Fig. 3. Absolute confusion degree of emotion. The darker/lighter the block, the longer/shorter distance from the targeting emotion to other emotions, the lower/higher the absolute confusion value ($A^g$). Anger in text is the most unlikely emotion to be confused with others in comment, love in text is the most unlikely emotion to be confused with others in news body, and fear in text is the most unlikely emotion to be confused with others in news title.

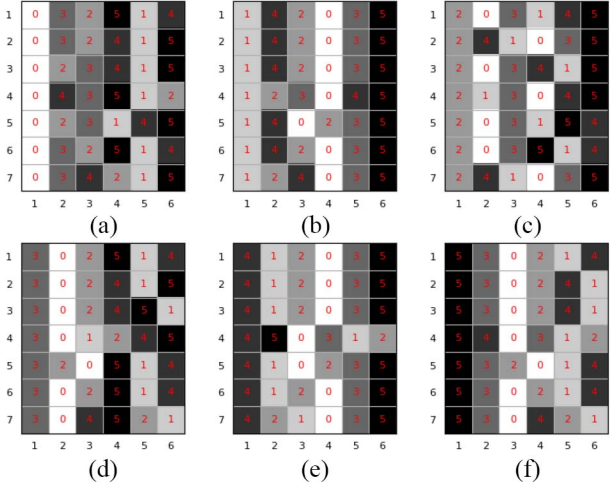

Fig. 4. The sequence matrix of comment. 0: love, 1: fear, 2: joy, 3: sadness, 4: surprise and 5: anger. *x* axes means the sequence from one to six. *y* axes denote seven different points of views. The confusion degree decreases along with the *x*-coordinate from two to six. The second and the sixth column refer to the emotion of maximal and minimal confusion degree with the emotion in the first column, respectively. Taking (b) comment_fear for example. The first row of this matrix shows that 4 surprise is most likely to be confused (column 2) with 1 fear from the first point of view. Conversely, 5 anger is least likely (column 6) to be confused with 1 fear from the first point of view.

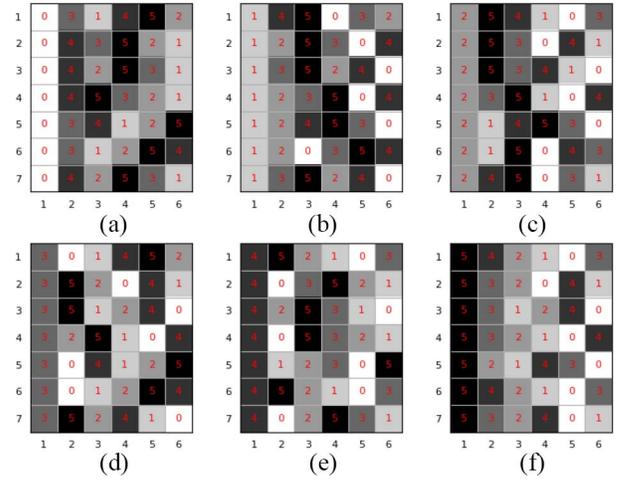

Fig. 5. The sequence matrix of news body. Refer to the title of Fig. 4.

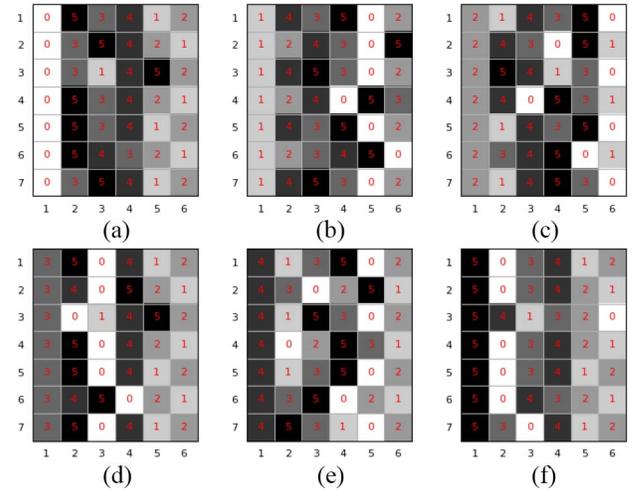

Fig. 6. The sequence matrix of news title. Refer to the title of Fig. 4.

### B. Confusion of Emotion Result

Considering the three kinds of features (explicit expression, implicit expression, and character) and the two models, there are six (three times two) analytic logic for each dataset. The average of the six-emotion correlation matrix is also considered. That means one dataset is evaluated from seven perspectives. The six categories of emotions are adopted, that is, love, fear, joy, sadness, surprise, and anger. Specifically, $m_0$ and $m_1$ both equal six. $C$, $B$, and $T$ indicate the comment, news body, and title datasets, which means $\mathbf{X}_1 = (\mathbf{C}_1, \mathbf{B}_1, \mathbf{T}_1)$, and $\mathbf{x}_1^{(k)} = (\mathbf{c}_1^{(k)}, \mathbf{b}_1^{(k)}, \mathbf{t}_1^{(k)})$.

The first four principal component vectors are adopted, as the variance contribution of which is greater than 85% and approaches 90%. Then, (7)–(9) are applied for the emotional features and absolute differences of the three datasets.

*1) Absolute Confusion Degree:* The absolute confusion degree is obtained through (10). The result is shown in Fig. 3. In the comments dataset, the absolute confusion degree of anger is the lowest among the six kinds of emotions while that of love is the highest. It follows that anger is unlikely to be confused with other emotions. Absolute confusion degree of emotion love is proportional to two elements, for example, the probability that love is mistaken as other emotions and other emotions are misjudged as love. The increase of either element would enhance the absolute confusion degree of emotion love. In the news body dataset, the absolute confusion degree of love is the lowest while that of joy is the highest. In the news title dataset, fear has the lowest absolute confusion degree while surprise has the highest one.

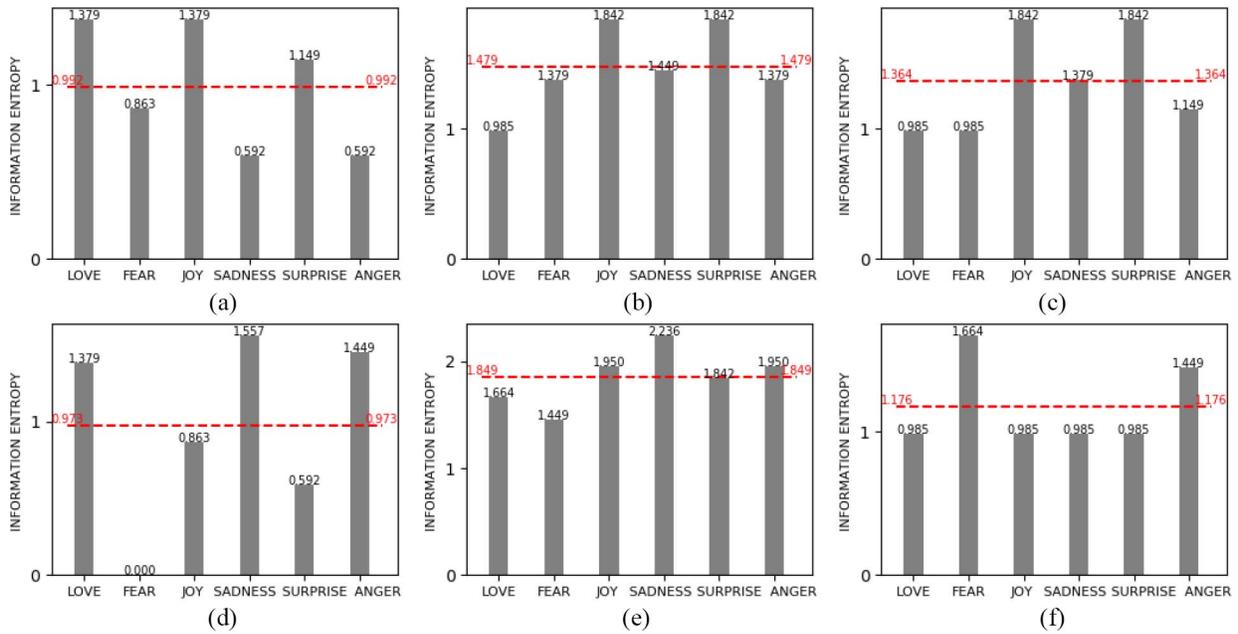

Fig. 7. Information entropy of the second and last column in the sequence matrices $\mathbf{X}_4^g$. The red dashed line represents the mean of the information entropy.

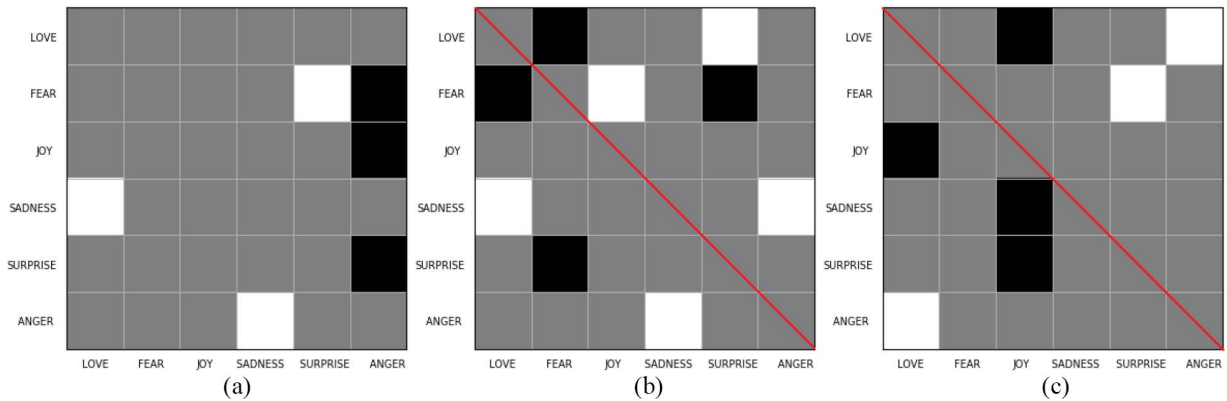

Fig. 8. Relative confusion degree of emotion. The white blocks represent the maximal confusion degree while the black blocks indicate the minimal degree. The gray blocks denote that the two emotions are neither most nor least confused or neglected according to information entropy which is shown in Fig. 7.

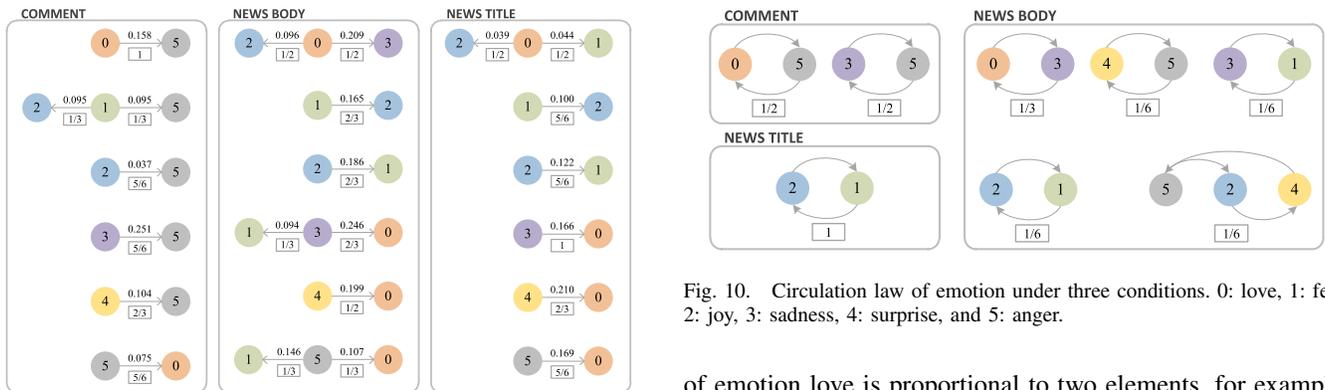

Fig. 9. Misjudgment law of emotion. 0: love, 1: fear, 2: joy, 3: sadness, 4: surprise, and 5: anger.

Fig. 10. Circulation law of emotion under three conditions. 0: love, 1: fear, 2: joy, 3: sadness, 4: surprise, and 5: anger.

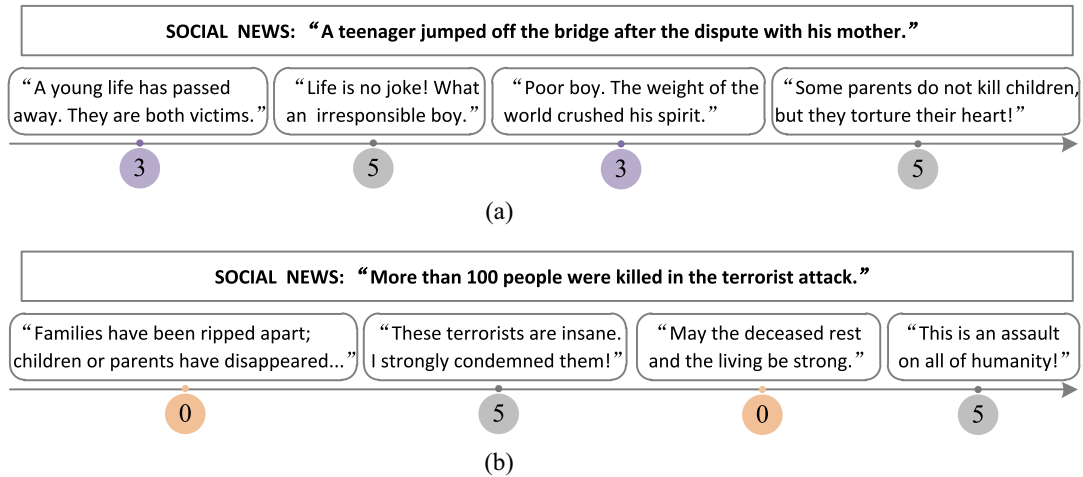

Fig. 11. Comment circulations of social news. 0: love, 3: sadness, and 5: anger. (a) *Sadness–anger* circulation in the comments. (b) *Love–anger* circulation in the comments.

*2) Relative Confusion Degree:* According to (11), the sequence matrix of the three datasets is shown in Figs. 4–6. In Fig. 4, the sixth column elements of fear-centered matrix are consistent, which means anger is least likely to be confused with fear in the comments from seven different perspectives. On the contrary, in Fig. 5, the sixth column of the fear-centered matrix is highly unordered, the reliability of which is low.

Fig. 7 shows the information entropy of the second and the last column which correspond to the emotion of the maximal and the minimal confusion degree, respectively. The mean of the information entropy decreases in the order: news body, title, and comment, stemming from the complexity of the long text in the news body and the difference in objective and subjective texts. The columns of the sequence matrix with information entropy below the average value exhibit more reliable results, which are further analyzed. For example, in the comments, the emotions with the maximal confusion degree with fear, sadness, and anger are extracted. The emotions of the minimal confusion with fear, joy, and surprise attract our attention. The confusion law is shown in Fig. 8.

The relative confusion degree differs for comment, news body, and title, due to different traits of datasets. In the comments, anger is the emotion of the minimal confusion degree with fear, joy, and surprise. Conversely, surprise, love, and sadness have the maximal confusion degree with fear, sadness, and anger respectively. In the news bodies, the emotion confusion law is more dispersed. Love and fear are most unlikely to be confused with each other as the two emotions are on the diagonal symmetry. Sadness and anger are easy to be confused mutually. The long text of news body, which carries multiple information, contributes to the complexity of the emotional confusion. In the news titles, the models are unlikely to confuse joy and love while they tend to confuse anger and love. Joy has the minimal confusion degree with love, sadness, and surprise. Surprise is the emotion that is most likely to be confused with fear.

For the three datasets, anger and love are closely correlated in either a direct way (i.e., in the news titles) or an indirect way (i.e., in the comments and news bodies). In the news titles, the two emotions tend to be confused mutually. In the comments and news bodies, love is the emotion that has the maximal confusion degree with sadness, and sadness is the emotion that has the maximal confusion degree with anger. Anger and love are therefore associated indirectly. For example, the news title "more than 100 people were killed in the terrorist attack," which may arouse the mixture public emotions of anger and love.

### C. Evolution of Emotion Result

*1) Misjudgment of Emotion:* According to (13) in Section V-A, the top misunderstanding emotion pairs approved by two or more emotion correlation matrices are illustrated in Fig. 9. For example, in the comments, one-third of the matrices are likely to misinterpret fear as anger, the average probability of which is 0.095. One-third of the matrices recognize fear as joy with the same probability.

Overall, comments can be easily misjudged as anger, especially the texts that involve sadness and love. Conversely, anger is unlikely to be mistaken in the comments. In news body and title, the models are likely to be confused between fear and joy. The texts that cause sadness, surprise, and anger are easy to be recognized as love incorrectly.

*2) Circulation of Emotion:* The three conditions mentioned in Section V-B are considered for the most-likely traces in the evolution of emotion. Eight emotion shift steps are employed for the three conditions considering the validity and efficiency of computation. The most-likely traces include the phenomenon of emotion circulation under the above three conditions. Moreover, the emotion circulations that are extracted from an emotion correlation matrix $\mathbf{x}_1^{(k)}$ are the same, even though the original condition differs. The circulation law of emotion under three conditions, by (14)–(16), can therefore be concluded as Fig. 10.

In the comments, half the traces of the maximal probability contain the circulation of love and anger, sadness, and anger. Take love and anger as an example, in computational terms, the circulations represent that $\text{prob}_1(0, 5) * \text{prob}_1(5, 0)$ are larger than other probability products. In emotional terms, comments that cause love and anger are easy to be confused mutually.

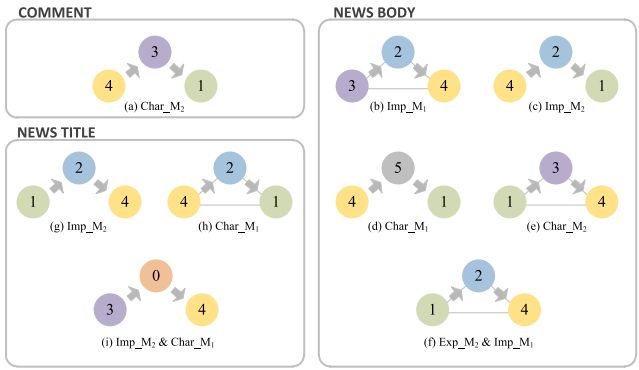

Fig. 12. The shortest path of two-steps shift rather than one-step. 0: love, 1: fear, 2: joy, 3: sadness, 4: surprise, and 5: anger.

The circulations result from the higher probability that love and sadness are misinterpreted as anger in Section VI-C1. Two cases are shown in Fig. 11. In the objective texts, especially in the news titles, there are the circulations of fear and joy, which satisfies the analysis in Section VI-C1.

*3) Shortest Path of Emotion Transfer:* Equation (17) in Section V-C is applied for the shortest path between two emotions. The results reveal that most shortest paths are one step. The shortest path can be two steps when $\text{prob}_1(s_{\text{ini}}, s_{ult})$ is equal or near to zero shown in Fig. 12.

Let Imp_$M_2$ represent the emotion correlation matrix obtained through the analysis of implicit expression by the second model. Take the news title as an example. As $\text{prob}_1(1, 4)$ is equal to zero, the shortest path between fear and surprise in Imp_$M_2$ is established with joy, formulated as $\text{trace}_2(1, 4) = 1 \rightarrow 2 \rightarrow 4$. Besides, as $\text{prob}_1(4, 1)$ is close to zero in Char_$M_1$, the shortest path is $\text{trace}_2(4, 1) = 4 \rightarrow 2 \rightarrow 1$. The horizontal lines represent that it is possible to misinterpret surprise as fear, that is, $\text{prob}_1(4, 1)$ is close to zero rather than equaling to zero. Most matrices support that the shortest path is one step. There exists the shortest path of two-shift steps between two emotions under some circumstances.

## VII. CONCLUSION

Mining the emotion correlation is significant for tracking emotional development. The complexity of emotion and event make emotion recognition hard for both subjective and objective information. The contributions of this article are concluded as follows.
1) This article mines the correlation of emotions based on the emotion recognition result of state-of-the-art deep learning models.
2) The errors caused by the dataset and models are cut down by designing three kinds of features and two deep neural-network models.
3) The emotion correlation is mined through an emotion confusion law, which is undirected, and an emotion evolution law, which is directed.

Experiments on subjective–objective and long-short Web news texts are conducted. There are several promising discoveries of emotion correlations.

1) Emotion correlation differs for different types of datasets.
2) In subjective comments, the absolute confusion degree of anger is low for the reason that anger is unlikely to be mistaken, even though all other emotions are easy to be recognized as anger. In objective texts, some emotions tend to be misinterpreted as love.
3) Comments arouse the emotion circulations of love and anger, and sadness and anger. In news body and title, fear and joy are the repeated emotions.
4) For specific datasets, some emotions are hard to be confused through one shift, but with higher probability to be confused through two shifts. This phenomenon conforms to the diversity of public emotion. The emotion correlation (confusion and evolution law) is potentially applicable to a wide variety of situations and events, such as public sentiment analysis, social media communication, and HCI.

This article employs Sina News in Chinese as a case study. Note that diverse culture leads to different results. However, the methods presented for exploring emotion correlation are applicable in various languages and domains.

## APPENDIX A
## EMOTION CLASSIFICATION MODEL

Here, two deep neural-network models, CNN-LSTM2 (M1) and CNN-LSTM2-STACK (M2), are employed for emotion recognition. In both models, the length of an input text can be either short or long. The output of the models is one of the six kinds of emotions, that is, love, joy, anger, sadness, fear, and surprise. The calculation process can be divided into three parts. M1 is constructed with Part I and Part II. M2 is constructed by adding an additional Part III to M1.

### A. Part I (Feature Processing)

Part I focuses on feature processing which transforms the original features into dense vector information. There are four operations in this part: 1) vector lookup; 2) window sliding; 3) convolutional calculation; and 4) ReLUs activation.

Let the input be denoted as $w_i^{(j)}$, which means the $i$th feature of the $j$th sample text. The $j$th sample text is indicated as $[w_1^{(j)}, w_2^{(j)}, \ldots, w_N^{(j)}]$, where the text is padded by "none" to length of $N$. Here, none is the reserved symbol in the vocabulary. For instance, if the first sample text is "I like small cat" and $N = 5$, then $w_1^{(1)} = $ "I," $w_2^{(1)} = $ "like," $w_3^{(1)} = $ "small," $w_4^{(1)} = $ "cat," and $w_5^{(1)} = $ "none."

The first operation, vector lookup, searches the embedding representation $z_i^{(j)}$ of the corresponding input feature $w_i^{(j)}$, which is formulated as

$$z_i^{(j)} = \text{LOOKUP}\left(w_i^{(j)}\right). \tag{18}$$

The second operation, window sliding, packages a target input feature and its context together after none padding. Specifically, the window size is set to 5, and $w_0^{(j)} = w_{-1}^{(j)} = $ "none." Then, in the third operation, a convolutional

layer is applied

$$h_{i,1}^{(j)} = \text{CNN}\left(\left[z_{i-2}^{(j)}, z_{i-1}^{(j)}, z_i^{(j)}, z_{i+1}^{(j)}, z_{i+2}^{(j)}\right]\right) \quad (19)$$

where $h_{i,1}^{(j)}$ is the first hidden layer. The introduction of hidden layers contributes to no-linear behavior exhibited by the network. $[\cdot]$ means embedding concatenation. In the last operation, a ReLU activation layer is added. ReLU is one of the most commonly used activation functions, which has strengths for vanishing gradient problem, higher computation, and convergence speed

$$h_{i,2}^{(j)} = \text{ReLU}\left(h_{i,1}^{(j)}\right) \approx \log\left(h_{i,1}^{(j)}\right) \quad (20)$$

where $h_{i,2}^{(j)}$ is the second hidden layer of the $i$th feature of the $j$th sample. $h_{i,2}^{(j)}$ acts as the input for Part II.

### B. Part II (Emotion Calculation)

Part II focuses on the emotion calculation after feature processing in Part I. There are five operations in this part, that is, LSTM calculation, dropout operation, average calculation, fully connected layer, and softmax. RNNs are powerful for sequence processing tasks, such as text classification. First, the output of Part I ($h_{i,2}^{(j)}$) is fed into a two-layer LSTM component, the outputs of which are represented as $h_{i,3}^{(j)}$ and $h_{i,4}^{(j)}$, respectively. After that, the dropout operation is applied to prevent overfitting

$$h_{i,3}^{(j)} = \text{LSTM}\left(h_{i,2}^{(j)}\right) \quad (21)$$
$$h_{i,4}^{(j)} = \text{LSTM}\left(h_{i,3}^{(j)}\right) \quad (22)$$
$$h_{i,5}^{(j)} = \text{DROPOUT}\left(h_{i,4}^{(j)}\right) \quad (23)$$

where $i$ is the index of the text sequence, which is an integer ranging from 1 to $N$.

In practice, even though the texts have been padded to the same length of $N$, their actual lengths still vary. To settle this problem, $ms_i^{(j)} \in \{0, 1\}$ is defined as the mask. The sequence data are combined to a fixed-length vector

$$h_{\_,6}^{(j)} = \frac{1}{N} \sum_i \left(h_{i,5}^{(j)} \cdot ms_i^{(j)}\right). \quad (24)$$

If the $i$th feature of the $j$th sample text is valid (i.e., not none), then $ms_i^{(j)} = 1$. Otherwise, $ms_i^{(j)} = 0$. Note that $h_{\_,6}^{(j)}$ denotes the sixth hidden layer. The last two steps are a fully connected layer and a softmax layer

$$h_{\_,7}^{(j)} = \text{LINEAR}\left(h_{\_,6}^{(j)}\right) = W^T h_{\_,6}^{(j)} + b \quad (25)$$
$$\hat{y}^{(j)} = \text{SOFTMAX}\left(h_{\_,7}^{(j)}\right) \quad (26)$$

where $W$ and $b$ are the weight matrix and bias, respectively. $l$ is the emotion index, which refers to one of the six emotions. $\hat{y}^{(j)}$ means the predicted probability distribution of six categories of emotions for the $j$th input data. Corresponding to the model-predicted label ($\hat{Y}$), $\hat{y}_l^{(j)}$ denotes the predicted probability of emotion $l$ for the $j$th input data. Corresponding to the model input label (Y), $y_l^{(j)}$ indicates the public-annotated label (i.e., emotion $l$) for the $j$th input data, the value of which is one in the case study. The loss describes the distance between probability of the predicted label and the input label, which is calculated through cross entropy function

$$\text{loss} = -\sum_{l,j} y_l^{(j)} \log\left(\hat{y}_l^{(j)}\right). \quad (27)$$

TABLE I
TESTING AND TRAINING DATA ILLUSTRATION ON THREE DATASETS, COMMENT, NEWS BODY, AND NEWS TITLE

| Comment | Emotion | Love | Fear | Joy | Sadness |
|---|---|---|---|---|---|
| | Training | 29658 | 2098 | 15426 | 9238 |
| | Testing | 7398 | 520 | 3848 | 2304 |
| | Emotion | Surprise | Anger | Total | |
| | Training | 13283 | 51610 | 121313 | |
| | Testing | 3311 | 12882 | 30263 | |
| News Body | Emotion | Love | Fear | Joy | Sadness |
| | Training | 6788 | 5240 | 5178 | 1457 |
| | Testing | 1698 | 1309 | 1295 | 364 |
| | Emotion | Surprise | Anger | Total | |
| | Training | 323 | 578 | 19564 | |
| | Testing | 82 | 144 | 4892 | |
| News Title | Emotion | Love | Fear | Joy | Sadness |
| | Training | 6717 | 5211 | 5159 | 1451 |
| | Testing | 1680 | 1302 | 1290 | 364 |
| | Emotion | Surprise | Anger | Total | |
| | Training | 272 | 571 | 19391 | |
| | Testing | 68 | 142 | 4846 | |

### C. Part III (Original Feature Attention)

Part I and Part II construct the model CNN-LSTM2. However, with the neural network going deep, the backward fine-tuning process in CNN-LSTM2 becomes weak, and the vanishing gradient problem occurs. To solve this problem, a second model CNN-LSTM2-STACK is constructed by associating CNN-LSTM2 with Part III. This part links the input feature embedding $z_i^{(j)}$ to the layer $h_{\_,6}^{(j)}$ through linear and sigmoid operations

$$h_{i,8}^{(j)} = \text{LINEAR}\left(z_i^{(j)}\right) \quad (28)$$
$$h_{i,9}^{(j)} = \text{SIGMOID}\left(h_{i,9}^{(j)}\right) \quad (29)$$

Then, the layer $h_{\_,6}^{(j)}$ is adjusted by $h_{i,9}^{(j)}$, that is, the inputs of the sixth hidden layer change to both $h_{i,5}^{(j)}$ and $h_{i,9}^{(j)}$

$$h_{\_,6}^{(j)} = \frac{1}{N} \sum_i \left(h_{i,5}^{(j)} \cdot ms_i^{(j)}\right) + \frac{1}{N} \sum_i \left(h_{i,9}^{(j)} \cdot ms_i^{(j)}\right). \quad (30)$$

The other operations in Part II remain the same. Part III aims to emphasize the impact of the input feature embedding on the emotion calculation result. In other words, by stacking Part III, the network pays more attention to the original feature information.

The two models CNN-LSTM2 and CNN-LSTM2-STACK use deep neural-network methods to calculate text emotion.

TABLE II
ACCURACY OF TWO MODELS WITH THREE FEATURES ON THE TESTING DATA OF DATASET COMMENT, NEWS BODY, AND NEWS TITLE. MODEL M1 WORKS WITH PART I AND PART II AS DESCRIBED IN APPENDIX A. MODEL M2 WORKS WITH PART I, PART II, AND PART III AS DESCRIBED IN APPENDIX A

|  | Model | Explicit | Implicit | Character |
| --- | --- | --- | --- | --- |
| Comment | M1 | 0.850 | 0.822 | 0.844 |
|  | M2 | 0.829 | 0.761 | 0.815 |
| News Body | M1 | 0.812 | 0.796 | 0.620 |
|  | M2 | 0.825 | 0.791 | 0.553 |
| News Title | M1 | 0.800 | 0.778 | 0.796 |
|  | M2 | 0.805 | 0.812 | 0.820 |

The detailed description are demonstrated by Wang *et al.* [62]. In our model design, both long and short texts can act as inputs. Three parts are introduced, including feature processing, emotion calculation, and original feature attention. The details are shown in Fig. 2. The emotion calculation results of the two models with three text features support the inter emotion correlation analysis.

## APPENDIX B
## DATASET AND MODEL ACCURACY

See Tables I and II.

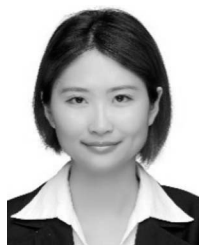

**Xinzhi Wang** received the Ph.D. degree from Tsinghua University, Beijing, China, in 2019.

She is currently an Assistant Professor with the School of Computer Engineering and Science, Shanghai University, Shanghai, China. She visited Carnegie Mellon University, Pittsburgh, PA, USA, from 2017 to 2018. Her recent research focuses on natural language and image processing, including sentiment analysis, AI transparency, agent action planning, and agent action intervention.

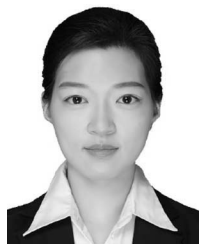

**Luyao Kou** received the bachelor's degree from the School of Resources and Safety Engineering, China University of Mining and Technology, Beijing, China, in 2018. She is currently pursuing the Ph.D. degree with the Institute of Public Safety Research, Department of Engineering Physics, Tsinghua University, Beijing.

Her main research field includes sentiment analysis, building system and safety, and emergency management.

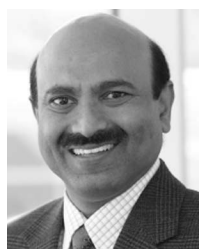

**Vijayan Sugumaran** received the Ph.D. degree in information technology from George Mason University, Fairfax, VA, USA, in 1993.

He is a Distinguished Professor of management information systems, the Chair of the Department of Decision and Information Sciences, and the Co-Director of the Center for Data Science and Big Data Analytics, Oakland University, Rochester, MI, USA. He has published over 250 peer-reviewed articles in journals, conferences, and books, such as *Information Systems Research*, the *ACM Transactions on Database Systems*, the IEEE TRANSACTIONS ON BIG DATA, and the IEEE TRANSACTIONS ON EDUCATION. He has edited 20 books and serves on the editorial board of eight journals. His research interests include big data management and analytics, ontologies and semantic Web, intelligent agent, and multiagent systems.

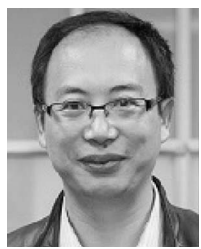

**Xiangfeng Luo** received the Ph.D. degree from Hefei University of Technology, Hefei, China, in 2003.

He is a Professor with the School of Computer Engineering and Science, Shanghai University, Shanghai, China. He has visited Purdue University, West Lafayette, IN, USA, as a Visiting Professor. He has authored or coauthored more than 100 publications and his publications have appeared in the IEEE TRANSACTIONS ON SYSTEMS, MAN, AND CYBERNETICS—PART C: APPLICATIONS AND REVIEWS and the IEEE TRANSACTIONS ON AUTOMATION SCIENCE AND ENGINEERING. His main research interests include Web wisdom, cognitive informatics, and text understanding.

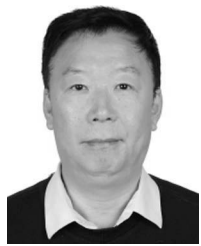

**Hui Zhang** received the Ph.D. degree from the Polytechnic Institute of New York University, New York, NY, USA, in 1994.

He is the Vice Director of the Institute of Public Safety Research, Tsinghua University, Beijing, China. His recent research mostly focuses on public safety and emergency management, including preparedness and response to emergencies, sentiment analysis, decision-making support, applications of social media, and big data in emergency management.

Dr. Zhang received the National Science Foundation CAREER Award (USA) in 1999 when he was at SUNY, Stony Brook, NY, USA, and the Changjiang Scholar (China) in 2008.